\documentclass[journal,twocolumn,transmag]{IEEEtran}
\usepackage{caption}
\usepackage{times}
\usepackage{epsfig}
\usepackage{graphicx}
\usepackage{amsmath}
\usepackage{amssymb}

\usepackage{bbding}      

\usepackage{multirow}
\usepackage{subfigure}
\usepackage{ulem}
\usepackage{color} 
\usepackage{enumerate}

\usepackage[backref]{hyperref}
\usepackage{makecell}

\newcommand{\ie}{\textit{i.e.} }

\ifCLASSINFOpdf
\else
\fi
\begin{document}
%

\title{Temporal Action Localization with Multi-temporal Scales}

\author{Zan Gao, Member, IEEE, Xinglei Cui, Tao Zhuo, Zhiyong Cheng, An-An Liu, Senior Member, IEEE, \\ Meng Wang, IEEE Fellow, and Shenyong Chen, Senior Member, IEEE, IET Fellow}



%




\maketitle
\begin{abstract}
Abstract -Temporal action localization plays an important role in video analysis, which aims to localize and classify actions in untrimmed videos. The previous methods often predict actions on a feature space of a single-temporal scale. However, the temporal features of a low-level scale lack enough semantics for action classification while a high-level scale cannot provide rich details of the action boundaries. To address this issue, we propose to predict actions on a feature space of multi-temporal scales. Specifically, we use refined feature pyramids of different scales to pass semantics from high-level scales to low-level scales. Besides, to establish the long temporal scale of the entire video, we use a spatial-temporal transformer encoder to capture the long-range dependencies of video frames. Then the refined features with long-range dependencies are fed into a classifier for the coarse action prediction.
Finally, to further improve the prediction accuracy, we propose to use a frame-level self  attention module to refine the classification and boundaries of each action instance. Extensive experiments show that the proposed method can outperform state-of-the-art approaches on the THUMOS14 dataset and achieves comparable performance on the ActivityNet1.3 dataset. Compared with A2Net (TIP20, Avg\{0.3:0.7\}),  Sub-Action (CSVT2022, Avg\{0.1:0.5\}), and AFSD (CVPR21, Avg\{0.3:0.7\}) on the THUMOS14 dataset, the proposed method can achieve improvements of 12.6\%, 17.4\% and 2.2\%, respectively \footnote{Manuscript received July 7th, 2022; This work was supported in part by the National Natural Science Foundation of China (No.61872270, No.62020106004, No.92048301, No.61572357).  Young creative team in universities of Shandong Province (No.2020KJN012), Jinan 20 projects in universities (No.2020GXRC0404, No.2018GXRC014). 

Z. Gao, X.L Cui (Co-first author), T. Zhuo (Corresponding author) and Z.Y Cheng are with Shandong Artificial Intelligence Institute, Qilu University of Technology (Shandong Academy of Sciences), Jinan, 250014, P.R China. 

A.A Liu is with the School of Electrical and Information Engineering, Tianjin University, Tianjin, 300072, China. 

M. Wang is with the school of Computer Science and Information Engineering, Hefei University of Technology, Hefei, 230009, P.R China. 

S.Y Chen and Z.Gao are with Key Laboratory of Computer Vision and System, Ministry of Education, Tianjin University of Technology, Tianjin, 300384, P.R China.
}.  

\end{abstract}


\maketitle

\IEEEdisplaynontitleabstractindextext

%
\IEEEpeerreviewmaketitle

\section{Introduction}
\begin{figure*}
\centering 
\includegraphics[width=0.9\textwidth,height=0.32\linewidth]{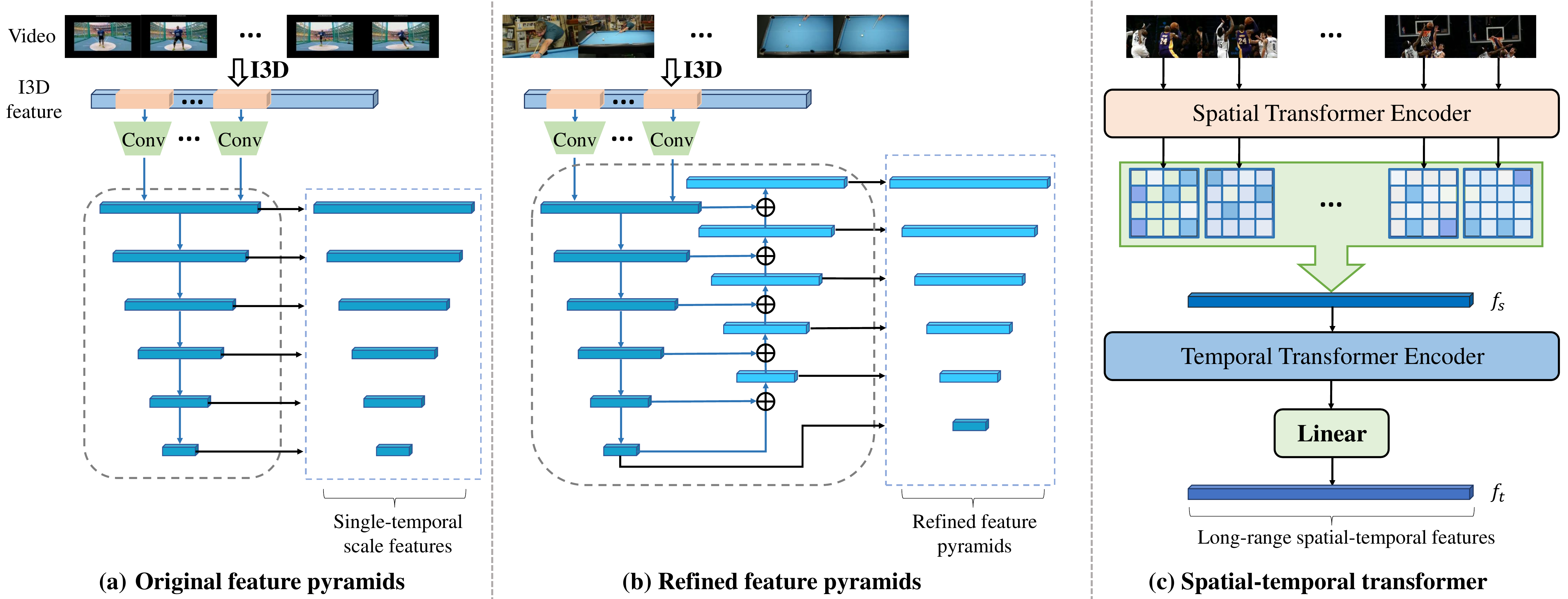}

\caption{ (a) Original feature pyramids with the single-temporal scale. (b) Refined feature pyramids with multi-temporal scales pass semantics from high-level to low-level scales. (c) Spatial-temporal transformer for long-range dependencies.
}
\label{fig_introductin}
\end{figure*}
%
%
%
%

In recent years, with the emergence of a large number of Internet videos, Temporal Action Localization (TAL) has attracted lots of attention in academia and industry \cite{Wang2022Exploring, Stoian2016Fast}. As an important branch of video understanding, the goal of TAL is to locate the start and end of each action instance in untrimmed videos and predict the category of actions.

According to different processing strategies, recent TAL methods can be roughly divided into three categories, anchor-based, actionness-guided, and anchor-free methods. The anchor-based methods \cite{Gao2017TURN, Chao2018Rethinking,Xu2017R-C3D} generate a set of action proposals with the pre-defined anchors at different temporal scales. Then the actions are classified and the boundaries are regressed. However, due to the fixed pre-defined anchors, the anchor-based methods are not flexible enough for various action categories. Besides, the anchor-based methods are very sensitive
to some hyper-parameters. For flexible TAL, the actionness-guided approaches \cite{Lin2018BSN, Lin2019BMN, Xu2020G-TAD, Zeng2019PGCN, Su2021BSN++, Lin2020DBG, Shou2017CDC} focus on predicting the confidence scores of the start probability, end probability, and duration of action, and then combining them into proposals. Unlike anchor-based methods, actionness-guided approaches do not require pre-defined anchors to generate action proposals. However, such a strategy requires an extra model for action classification, and its computational cost is relatively expensive. Recently, for efficient TAL in videos, the anchor-free strategy \cite{Ou2022SRFNet, Lin2021AFSD} only needs to generate a proposal at each temporal position, which is formed by combining the regions from the current position to the start position and end position, and it also does not require the pre-defined anchors. Compared to the actionness-guided methods, anchor-free based methods do not enumerate boundaries, avoiding the redundancy of proposals and reducing the amount of computation.

In order to localize and classify actions with different temporal scales, the majority of existing anchor-free techniques often predict actions on the feature space of a single-temporal scale, \ie feature space of each pyramid layer individually, as illustrated in Figure \ref{fig_introductin}.a. However, the features of a low-level temporal scale lack enough semantics for action classification while a high-level scale cannot provide rich details of action boundaries. As a result, it is difficult to consider both the semantics and boundaries for each action instance simultaneously.

In this paper, we propose a Temporal Action Localization method by using features of Multi-Temporal Scales, namely TAL-MTS. As illustrated in Figure \ref{fig_introductin}.b, to make the feature contain enough semantics and rich details of boundaries, we use the nearest neighbor linear interpolation to merge the semantics information of the high-level temporal features into the low-level. For the Refined Feature Pyramids (RFP), the semantics information from the high-level temporal scale can be passed to the low-level one. Then both the semantics and details of actions can be considered simultaneously. Furthermore, to establish the long temporal scale of the entire video, we propose to use a Spatial-Temporal Transformer (STT) encoder to capture the long-range dependencies of video frames. As illustrated in Figure \ref{fig_introductin}.c, the factorized encoder consists of two transformer encoders in series, a spatial transformer encoder models the latent representation of each video frame, and a temporal transformer encoder models the relationship between frames. Thus, it can represent the spatial-temporal context of the videos at a long-range scale. By combining the refined feature pyramids and long-range spatial-temporal features on the entire video, the proposed method is able to well predict actions with robust features of multi-temporal scales.

In addition, to further enhance the foreground information of video frames, we perform a patch operation on each video frame and use self-attention to extract the relationship between patches to obtain features with salient foreground information. It is able to effectively reduce the influence of background noises on action instances, and enhance the foreground action features inside videos. Extensive experiments show that TAL-MTS achieves $56.9\%$ on MAP@0.5 leading state-of-the-art by $1.8\%$ on THUMOS14 dataset, and achieves comparable results on ActivityNet1.3 dataset.

The main contributions of this paper are summarized as follows:
\begin{itemize}
\item We propose a novel temporal action localization with refined features of multi-temporal scales, namely TAL-MTS. Compared to previous approaches, the proposed method is able to provide sufficient semantics, rich details of boundaries, and long-range dependencies for robust temporal action localization.

\item We propose to refine the classification and boundaries of action instances with a frame-level self-attention module, which can reduce the noises caused by backgrounds.

\item Extensive experiments show that our model outperforms state-of-the-art methods on the THUMOS14 dataset and obtains comparable results on the ActivityNet-1.3 dataset.

\end{itemize}

The remainder of this paper is organized as follows: Section II introduces related work, and Section III details the proposed TAL-MTS method. In Section IV, the experimental settings are presented, and the results are analyzed on the THUMOS14 dataset and ActivityNet1.3 dataset. In Section V, we present the ablation study. Finally, Section VI presents the conclusions.

\section{Related Work}
The TAL has become a hot topic in the field of computer vision. More and more researchers have proposed novel methods and achieved good results. In this section, We will introduce these methods from
three aspects anchor-based Localization, anchor-free methods, actionness-guided Localization. At the same time, we will also introduce transformers in computer vision.

\textbf{Anchor-based Localization.} The Anchor-based methods rely on pre-defined anchors of different scales, which are divided into one-stage and two-stage methods. SSAD \cite{Lin2017SSAD} and GTAN \cite{Long2019GTAN} methods are one-stage methods. For the SSAD \cite{Lin2017SSAD} method, it utilizes a single-shot structure based on 1D convolution to generate anchors for TAL, and GTAN \cite{Long2019GTAN} uses 3D ConvNet to extract small segment-level features, and uses temporal Gaussian kernel to generate proposals with different temporal resolutions. About two-stage methods, R-C3D \cite{Xu2017R-C3D} and TALNet \cite{Chao2018Rethinking} are similar to the structure of \cite{Ren2017Faster}. R-C3D \cite{Xu2017R-C3D} proposes an end-to-end network that combines candidate segment generation and classification to learn features and accepts input of videos of any length. TALNet \cite{Chao2018Rethinking} expands the receptive field and extracts temporal context for features. Simultaneously, late fusion is used for two-stream architecture. Different from the above two methods, TURN \cite{Gao2017TURN} divides the video into equal-length units, and includes extracting unit-level features, classifying action instances, and regressing temporal boundaries. Unlike common anchor-based detection techniques, RCL \cite{Wang2022RCL} proposes to use continuous Anchoring representation to achieve high-quality action detection. Confidence scores are regressed from continuous anchor points, and their confidence scores are jointly determined by video features and temporal coordinates. Although these methods can achieve good results, they are not flexible enough and also generate large redundancy.

\textbf{Anchor-free methods.} The anchor-free methods do not require pre-defined anchors, and the action proposal is represented by the distances from the current position to the start position and the end position. Thus there is no need for a large number of hyperparameters, and the computational cost is relatively low. For example, CornerNet \cite{Law2018CornerNet} uses a convolutional network to generate two sets of heat maps to predict corners for different categories, one for the upper left corner and the other for the lower right corner. And it will also find the offset positioned by the corner to make the bounding box more accurate. In TAL tasks, SRF-Net \cite{Ou2022SRFNet} designs a Selective Receptive Field Convolution (SRFC) mechanism, which can adaptively adjust the size of the receptive field according to multiple scales of input information at each temporal localization. The AFSD \cite{Lin2021AFSD} framework consists of three modules: feature extraction, coarse prediction, and refined prediction. It first extracts the pyramid features from the video, then predicts the boundary and classification of the coarse proposal for each pyramid layer, and finally optimizes the boundaries and classification of each proposal through the refined prediction. In addition, A2Net \cite{Yang2020Revisiting} introduces an anchor-free method to solve the problem of too long or too short video sequence length, and it can use the complementary properties to handle temporal sequences of different lengths. In this paper, our method is based on the anchor-free strategy, and we use features of Multi-temporal scales for temporal action localization.

\textbf{Actionness-guided Localization.} This strategy generates proposals by predicting confidence scores for the start probability, end probability, and duration. Different from the anchor-based approach, actionness-guided localization is more flexible in handling action instances that are too long or too short. Earlier in this method, TAG \cite{Xiong2017TAG} and SSN \cite{Zhao2017SSN} are proposed. TAG scores the sampled snippets, judges whether it is an action or not, and combines the snippets that are actions into proposals. SSN is based on the TAG method to generate proposals and divides the proposal into three stages: start, end, and activity. Each stage performs a pooling operation and then judges whether the action is complete and the action is classified. Soon after, Lin et al. proposed BSN \cite{Lin2018BSN} , LGN \cite{Lin2019LGN}, BMN \cite{Lin2019BMN} and BSN++ \cite{Su2021BSN++}. BSN first locates the boundaries of temporal action segments and directly combines the boundaries.
Then, proposal-level features are extracted based on the sequence of action confidence scores. Based on BSN, LGN proposes a " local to global " approach to jointly learn local and global contexts to generate action proposals, locally locate the accurate proposal boundary and globally evaluate the reliable confidence score. Improved BSN framework to BMN, which densely evaluates the confidence scores of all possible temporal sequences by generating a one-dimensional boundary probability sequence and a two-dimensional BM confidence map. Based on the above frameworks, BSN++ uses a boundary complementary classifier to enrich the context information for boundary prediction, and it designs a proposal relationship module that uses channel-wise and position-wise global dependencies to model proposal-proposal relationships. In addition, PGCN \cite{Zeng2019PGCN} firstly uses graph convolutional networks to capture proposal-proposal relationships. Each proposal is represented as a node, and the two proposals are represented as an edge. Two types of relationships are used, one for capturing the contextual information of each proposal and the other for describing the association between different actions. Similar to PCGN, GTAD \cite{Xu2020G-TAD} uses 3 GCNext modules for feature extraction, gradually aggregating temporal information and multi-level semantics information. Then, the extracted features are fed into the SGAlign layer, and the localization module obtains the scores of sub-graphs and sorts them, and then gets the final result. Since this method considers all possible combinations of time positions, it requires lots of computational costs.

\textbf{Transformers in computer vision.} The transformer architecture was first proposed in the Natural Language Processing (NLP) task, based on the self-attention mechanism to capture the relationship between sequences. Recently, it has received extensive attention in the field of computer vision, and the self-attention \cite{Vaswani2017Attention} mechanism in the transformer plays an important role in extracting temporal information in videos. For object detection, DETR \cite{Carion2020DETR} uses object queries instead of Anchors as candidates. A pure transformer architecture may not be sufficient to model complex temporal dependencies for action detection, Therefore, MS-TCT \cite{Dai2022mstct} uses convolutions to promote multiple temporal scales of tokens, and to blend neighboring tokens imposing a temporal consistency. Then, infusing local information between tokens. In the TAL task, ATAG \cite{Chang2021ATAG} proposes an augmented transformer with an adaptive graph network that captures both long-time and local temporal context information. TAPG \cite{Wang2021TAPG} designs a unified temporal action proposal generation framework, which consists of a boundary transformer and a proposal transformer, which capture long-term temporal dependencies and enrich the relationship between proposals respectively to predict accurate boundary information and reliable confidence assessment. ViT \cite{Dosovitskiy2021ViT} directly divides the image into patches of the same size and applies the transformer to the image patch sequence, which can well accomplish the task of image classification. Inspired by ViT, we use RFS to enhance the foreground information of the video. The Factorised encoder model in ViViT \cite{Arnab2021ViViT} based on ViT consists of a spatial transformer encoder and a temporal transformer encoder, which models the interactions of different temporal information. Thus, We design STT to extract the long-range spatial-temporal information in the video to enhance the features.

\begin{figure*}
\centering 
\includegraphics[ width=0.9\linewidth,height=0.35\linewidth]{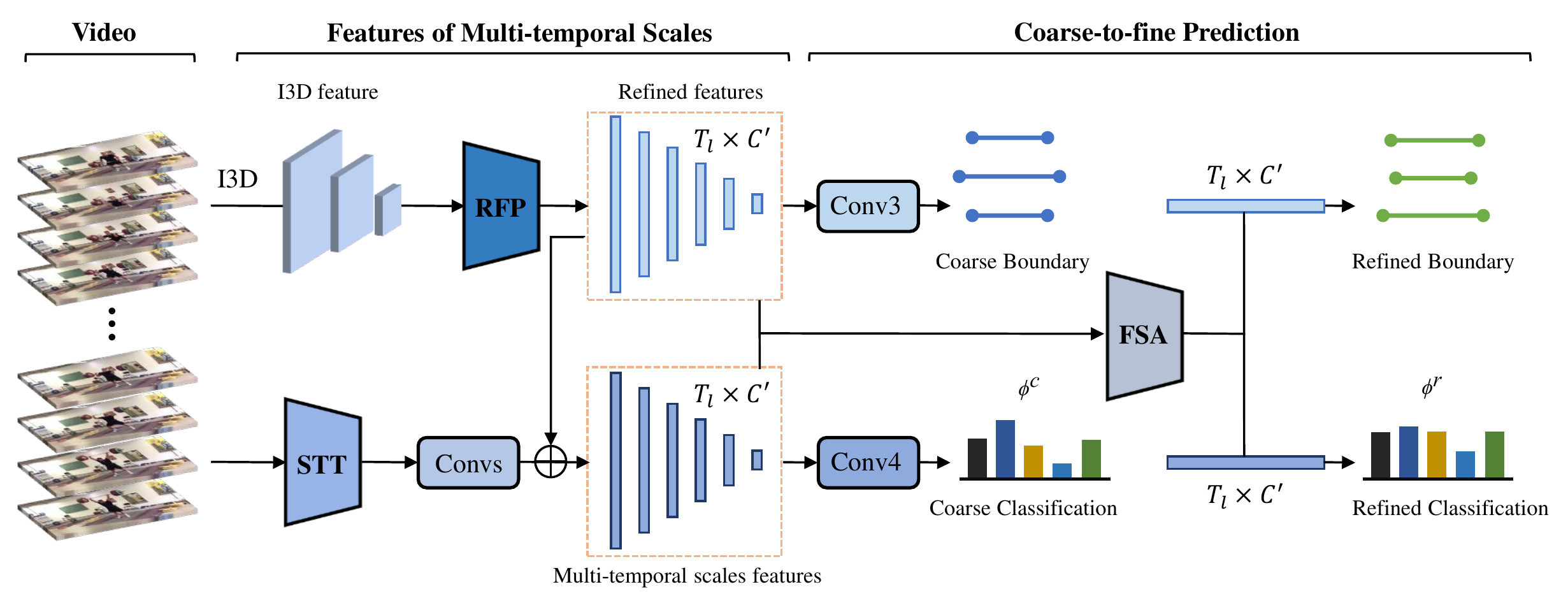}

\caption{An overview of the proposed TAL-MTS method. In this architecture, the I3D network is used as the feature extractor. The Refined Feature Pyramids (RFP) produces refined features with stronger semantics information at different temporal scales. The Spatial-Temporal Transformer (STT) excavates the spatial-temporal information of videos. The coarse-to-fine prediction is designed to minimize influence of video background noises, which  consists of an Frame-level Self Attention (FSA) module. Note that $\bigoplus$ is concatenation.
}
\label{fig_framework}
\end{figure*}

\section{Our Approach}
Suppose an untrimmed video $V=\left\{v_i\right\}_{i=1}^T$ consists of $T$ frames, the set of action instances  is represented by $X=\left\{t_{s,m},t_{e,m},\phi_m\right\}_{m=1}^M$, where $s_m$ and $e_m$ represent the start time and end time, repectively. $\phi_m$ and $M$ indicate the category of the $m$-th action instance and the number of action instances in the video, respectively. Our goal is to predict action segments with the start time, end time, and the corresponding action category. 

Unlike the previous methods \cite{Lin2021AFSD,Yang2020Revisiting,Ou2022SRFNet} that use the CNN model to exploit the features of single-temporal scale, we propose to use features of multi-temporal scales on videos. Based on an anchor-free TAL framework, our method is able to well locate and classify various actions.

\subsection{Overview}
The overview of the proposed method is illustrated in Figure \ref{fig_framework}. Given a video sequence, we first use I3D to extract features and then refine the feature pyramids by passing semantics from high-level scales to low-level ones, namely Refined Feature Pyramids (RFP) module.
Besides, we use a Spatial-Temporal Transformer (STT) module to capture the long-range dependencies of video frames. Then, by combining the features from RFP and STT, features of multi-temporal scales can be generated for the coarse action prediction. Finally, a Frame-level Self Attention (FSA) module is used to extract the salient foreground information, which further refines the action classification and boundaries (Section. 3.3). Next, we will describe the main modules of the proposed method.

\subsection{Features of Multi-temporal Scales}

The features of multi-temporal scales are generated from a Refined Feature Pyramids (RFP) module and a Spatial-Temporal Transformer (STT) module. Details of these two modules are as follows.

{\bf Refined Feature Pyramids (RFP).}
We use the I3D \cite{Carreira2017I3D} model pre-trained on the Kinetics dataset for both RGB frames and optical flows to obtain I3D features. For a video $V\in R^{C\times T\times H\times W}$, $C$ is the number of channels, $T$ is the temporal duration, $H$ and $W$ are the height and width of the video frame, respectively. The spatial-temporal features are denoted as $\mathbb{F}\in R^{C^{\prime}\times T^{\prime}\times H^\prime\times W^\prime}$ through the I3D network, where $C^\prime$, $T^\prime$, $H^\prime$ and $W^\prime$ also represent  channel, temporal duration, height and width respectively. Then, the spatial-temporal features are converted into a 1D feature space $\mathbb{F}_{st}\in R^{C^{\prime}\times T^{\prime}}$ through 3D convolutions, and by downsampling with 1D convolution layers, following \cite{Lin2021AFSD}, the feature pyramids with 6 different temporal scales are generated . Meanwhile,  the dimension is 512 for the feature pyramids. The first two temporal scales use 3D convolution downsampling with the kernel of [1, 6, 6] and [1, 3, 3] respectively. The remaining four temporal scales are downsampled using four identical 1D convolutions with kernel = 3 and stride = 2.

Inspired by the previous method \cite{Wang2021RGB}, we designed a RFP to compensate the semantics information to 1D feature sequences and expand the receptive field. To be specific, we add the semantics information of higher-level features to the lower-level features. Then, the features of different temporal scales with stronger semantics information can be obtained. Additionally, different from other similar structures, RFP uses nearest neighbor linear interpolation instead of convolution in the refined feature pyramids process. Thus, there is no need to update the parameters, which reduces the number of trainable parameters and speeds up the calculation. Based on RFP, refined feature pyramids with stronger semantics information $f\in R^{T_l\times C^{\prime}}$, $T_l\in\left\{2,4,8,16,32,64\right\}$ can be generated, where $T_i$ represents the different temporal spans. Moreover, a frame-level feature $f_{a}^c\in R^{T\times C^{\prime} }$ is generated by taking the feature of the lowest layer into a feature with temporal span $T$ adopting linear interpolate, which is used for further prediction refinement.

{\bf Spatial-Temporal Transformer (STT).} 
The long-range dependencies are important for the TAL task. Although we have got the refined feature pyramids at different temporal scales, the long-range spatial-temporal information is still lost because the downsampling using convolution has lower temporal receptive fields from video to spatial-temporal features. Therefore, we use an STT module to extract the long-range dependencies of video frames.

Inspired by ViViT \cite{Arnab2021ViViT}, we extract the long-range spatial-temporal information of videos with a factorised encoder, which consists of a spatial transformer encoder and a temporal transformer encoder in series. The spatial transformer encoder excavates the relationship within video frames from the same temporal index. Then, all the spatial feature outputs after temporal embedding are used as the input for the temporal transformer encoder. The temporal encoder connects the frames with different temporal indexes for capturing the temporal information, and we will obtain features with long-range spatial-temporal information from STT. Therefore, the long-range spatial-temporal information is supplemented to the output of the refined feature pyramids.

Specifically, our method employs Multi-Headed Self Attention (MSA) \cite{Vaswani2017Attention} in parallel, and Layer Normalisation (LN) \cite{Ba2016Layer} is applied before each MSA block. We first embed spatial position encoding for video $V=\left\{v_i\right\}_{i=1}^T$ and feed into the spatial transformer encoder and get a feature sequence $f_s$. Then, we embed temporal information to this feature sequence and take the feature into the temporal transformer encoder. Finally, feature $f_t \in R^{T_j\times C^{\prime}}$ 
with long-range spatial-temporal information at one temporal scale  is produced, where $T_j$ is temporal scale and $C^{\prime}$ is channel. Therefore, this process is denoted as follows:
\begin{equation}
\begin{aligned} 
& f_s={\rm MSA}\left({\rm LN}\left({v_i}\right)\right) \quad  i=1, \cdots,  T, \\
& f_t={\rm Linear}\left({\rm MSA}\left({\rm LN }\left(f_{s} \right)\right)\right),
\end{aligned}
\end{equation}
where $T$ represents the number of frames in the video and ${\rm Linear(\cdot)}$ represents the fully connected layer. In order to gain multi-scale features with long-range spatial-temporal information, we downsample $f_t$  through multiple 1D convolutions to produce feature sequences  $f_l \in R^{T_l\times C^{\prime}}$ at different temporal scales,  $T_l\in\left\{2,4,8,16,32,64\right\}$. Then, we followed by concatenating the feature sequences $f$ obtained by RFP and the feature sequences  $f_l$ to produce features
of multi-temporal scales $f_{mts}$, which is represented by:
\begin{equation}
f_{mts}=\left[f, f_l\right],
\end{equation} 
where $[\cdot]$ indicates the concatenation. We project features of multi-temporal scales and refined feature pyramids to $f_{cls}$ and $f_{loc}$ respectively in two 1D convolutions at each temporal scale, which is computed as follows:
\begin{equation}
\begin{aligned}
& f_{loc}={\rm Conv{1}}\left(f\right),\\
& f_{cls}={\rm Conv{2}}\left(f_{mts}\right),
\end{aligned}
\end{equation}
 $f_{loc}$ for localization  and $f_{cls}$ for classification of action instances.
Two 1D convolutions have the same kernel and stride. 
\subsection{Coarse-to-fine Prediction}
Based on the refined features of multi-temporal scales, coarse prediction results can be obtained with a classifier. 
To further improve the performance, we propose to use a Frame-level Self Attention (FSA) module for both the action category and boundary refinement. 

{\bf Coarse Prediction.}
Based on $f_{loc}$ and $f_{cls}$, we use the two 1D convolutions  to produce coarse start and end boundary distances $\left(d_n^s,d_n^e\right)$, and coarse class score $\phi_n^c$, respectively, which are computed as following:
\begin{equation}
\begin{aligned}
& d_n^s={\rm Conv3}\left(f_{loc}\right),\\
& d_n^e={\rm Conv3}\left(f_{loc}\right),\\
& \phi_n={\rm Conv4}\left(f_{cls}\right),
\end{aligned}
\end{equation}
where $n$ represents position in the different temporal scales and $d_n^s$ represents the distance from position $n$ to the start time, and $d_n^e$ represents the position from position $n$ to the end time. Then, the coarse boundaries $\left(t_{s,n}^{c}, t_{e,n}^{c}\right)$ can be inferred, where $t_{s,n}^{c}$ and $t_{e,n}^{c}$ represent the coarse starting and ending time of the corresponding $n$-th location in refined pyramid features and $n\in \{0, 1, 2, \dots, T_l-1\}$.
\begin{figure}
\centering 
\includegraphics[width=0.85\linewidth,height=0.45\textwidth]{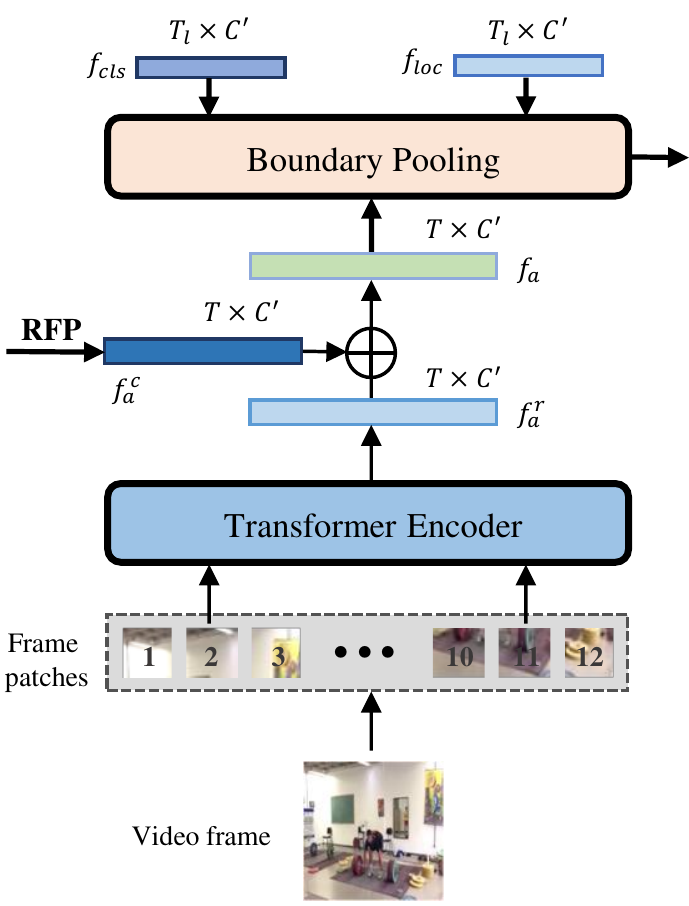}

\caption{Frame-level Self Attention (FSA) module. The entire video frame is divided into  frame patches and embedded with position encoding, and feed into the Transformer Encoder to generate the $f_{a}^r$. Then, combined with $f_{a}^c$ from RFP module to produce $f_{a}$. Finally, taking features $f_{loc}$ and $f_{cls}$ from coarse
prediction and  $f_{a}$   into the Boundary Pooling and obtaining features of fine-grained.
}
\label{fig_FSA}
\end{figure}

{\bf Frame-level Self Attention (FSA).} 
As the features of pyramid go deeper, the temporal dimension will become small and it is difficult to find accurate boundaries. Therefore, we use the frame-level feature for boundary pooling \cite{Lin2021AFSD}. Since frame-level features are derived from the lowest level of feature pyramids, they ignore the influence of internal background noises in each video frame. Thus, we use a  self-attention  \cite{Vaswani2017Attention} mechanism to minimize this influence. Inspired by ViT \cite{Dosovitskiy2021ViT}, we propose to use a Frame-level Self Attention (FSA) module to process the image of each frame, which maximizes the separation of foreground and background noises.

As illustrated in Figure \ref{fig_FSA}, we take all video frames as input of a transformer encoder and divide each video frame $v_i\in R^{H^{\prime}\times W^{\prime}\times C^{\prime}}$ into frame patches. Then, the relationship between patches in each frame can be built. Specifically, we divide the frame of $H^{\prime}\times W^{\prime}\times C^{\prime}$ into patches $v_{i, z}$  with the size of $D\times D$ where $z\in \{1, 2, \dots, Z\}$ and concatenate $Z$ patches into vectors that embedded patches positional encoding. Next, vectors will be fed into the transformer encoder, which 
consists of an MSA and an LN block. Finally, based on the outputs of the transformer encoder, we use the fully connected layer to get a frame-level feature to represent the foreground information, which is denoted as:
\begin{equation}
\begin{aligned}
& v_{i,z}={\rm TP}\left({v_i}\right)\quad i=1, \cdots,  T,\\
& f_{a}^r={\rm MSA}\left(v_{i, z}+{\rm PE}\left(v_{i, z}\right)\right), \quad z=1, \cdots,  Z,
\end{aligned}
\end{equation}
where ${\rm PE(\cdot)}$ represents patches position embedding for each video frame and ${\rm TP(\cdot)}$ stands for dividing video frames into patches. Frame-level features for refined prediction from $f_{a}^c$ generated by RFP and $f_{a}^r$ obtained by FSA. We concatenate $f_{a}^c$ and $f_{a}^r$ for refined frame-level feature $f_{a}$. Then, we take the features $f_{cls}$ and $f_{loc}$ in the coarse prediction and the frame-level feature $f_{a}$ in the fine prediction into boundary pooling \cite{Lin2021AFSD}. The features of fine-grained $ f_ {loc}^r$ and $f_{cls}^r$ are defined as follows:
\begin{equation}
\begin{aligned}
& f_{loc}^r={\rm BP}\left(f_{loc},f_{a}\right),\\
& f_{cls}^r={\rm BP}\left(f_{cls},f_{a}\right),
\end{aligned}
\end{equation}
where ${\rm BP}$ is a boundary pooling method \cite{Lin2021AFSD}. Fine-grained prediction for $f_{loc}^r$ and $f_{cls}^r$ are produced by two different 1D convolution layers respectively. Specifically, one convolution layer is used to predict the offsets $\left(\mathrm{\Delta}t_{s,n}^r,\mathrm{\Delta}t_{e,n}^r\right)$ for boundary regression, and the other is used to predict refined class score $\phi^r$. Finally, we add the offsets of boundaries $\left(\mathrm{\Delta}t_{s,n}^r,\mathrm{\Delta}t_{e,n}^r\right)$ to coarse boundary and get the refined boundaries $\left(t_{s,n}^r,t_{e,n}^r\right)$.

\subsection{Training and Inference}

{\bf Training.}
The set of action instances predicted by coarse prediction and refined prediction contains $N$ samples, which is larger than the maximum number of ground-truth action instances in the dataset. TAL-MTS uses multiple losses for the coarse boundary regression and classification and the binary cross-entropy loss is used for the probability of the proposal. The computation of our total loss function can be denoted by
\begin{equation}
L=L_{cls}^c+L_{cls}^r+\alpha\left(L_{loc}^c+L_{loc}^r\right)+\beta L_{bce} ,
\end{equation}
where $L$ is the total loss, $L_{cls}^c$ and $L_{cls}^r$ are the loss of the coarse classification and refined classification, respectively. $L_{loc}^c$ and $L_{loc}^r$ are the loss of the coarse boundary regression and refined boundary regression, respectively. $L_{bce}$ is the loss of binary cross-entropy \cite{Lin2021AFSD}. $\alpha$ and $\beta$ are hyper-parameters. For the coarse classification, focal loss \cite{Lin2020Focal} is applied as the constraint, because it can not only adjust the weight of positive and negative samples but also control the weight of difficult and easy classification samples, which is computed as follows:
\begin{equation}
L_{cls}^c=\frac{1}{N^c}\sum_{n=1}^{N^c}{\rm L_{focal}}\left(\phi_n^c, \phi_n\right),
\end{equation}
where $N^c$ is the number of the positive samples in the coarse process, it is regarded as a positive sample when it is located in the ground truth samples. $\phi_n^c$ is the coarse classification results and $\phi_n$ is the ground truth labels. For the refined classification, we use a focal loss as,
\begin{equation}
L_{cls}^r=\frac{1}{N^r}\sum_{n=1}^{N^r}{\rm L_{focal}}\left(\phi_n^r, \phi_n\right),
\end{equation}
where $N^r$ is the number of the positive samples when the coarse proposals have a tIoU higher than 0.5 with ground truth samples. $\phi_n^r$ is the refined classification results predicted and $\phi_n$ is the ground truth labels. We adopt GIoU loss \cite{Rezatofighi2019Generalized} as the constraint for coarse boundary regression, which is computed as follows:
\begin{equation}
L_{loc}^c=\frac{1}{N^c}\sum_{n=1}^{N^c}\left({1-{\rm GIoU}\left(\psi_n^c, \psi_n\right)}\right),
\end{equation}
where $\psi_n^c=\left(t_{s,n}^{c}, t_{e,n}^{c}\right)$ is the  coarse boundaries predicted by the coarse process, and $\psi_n=\left(t_{s,n}, t_{e,n}\right)$ is the corresponding ground truth. For the refined boundary regression, we use Smooth L1 loss \cite{Girshick2015Fast} as loss function and can be calculated as:
\begin{equation}
L_{loc}^r=\frac{1}{N^r}\sum_{n=1}^{N^r}\left({\rm {smooth_{L_1}}\left({\hat{\mathrm{\Delta}}}_n, \mathrm{\Delta}_n\right)}\right),
\end{equation}
where ${\hat{\mathrm{\Delta}}}_n=\left(\mathrm{\Delta}t_{s,n}^{c}, \mathrm{\Delta}t_{e,n}^{c}\right)$ is the offset between the coarse boundaries and the corresponding ground truth.  ${\mathrm{\Delta}}_n=\left(\mathrm{\Delta}t_{s,n}^{r}, \mathrm{\Delta}t_{e,n}^{r}\right)$ is the regression targets of our refined process. In addition, we use the binary cross-entropy loss to suppress proposals with low quality, which is defined as: 
\begin{equation}
L_{bce}=\frac{1}{N^r}\sum_{n=1}^{N^r}{\rm BCE}\left(\varepsilon_n, \frac{\left|\psi_n^r\cap\psi_n\right|}{\left|\psi_n^r\cup\mathrm{\psi}_n\right|}\right),
\end{equation}
where ${\rm BCE}$ is the binary cross-entropy loss. $\psi_n^r$ and $\psi_n$ are the refined boundaries and the corresponding ground truth, respectively.  $\varepsilon_n$ is the location.

{\bf Inference.}
In the inference stage, we use the coarse boundaries $\left(t_s^{c}, t_e^{c}\right)$, coarse classification results $\phi^{c}$, the offsets from refined process $\left(\mathrm{\Delta}t_s^{r}, \mathrm{\Delta}t_s^{r}\right)$, refined classification results $\phi^{r}$ and confidence scores $\varepsilon$ by our network. The final prediction for each cilp can be computed as: 
\begin{equation}
\begin{aligned}
& t_{s,n}^{p}=t_{s,n}^{c} + \frac{1}{2}d_n^{c}\mathrm{\Delta}t_{s,n}^{r},\\
& t_{e,n}^{p}=t_{e,n}^{c} + \frac{1}{2}d_n^{c}\mathrm{\Delta}t_{e,n}^{r},\\
& \phi_n^{p}=\frac{1}{2}\left(\phi_n^{c} + \phi_n^{r}\right)\varepsilon_n,
\end{aligned}
\end{equation}
where the $d_n^{c}=t_{e,n}^{c}-t_{s,n}^{c}$ . Finally, we adopt Soft-NMS \cite{Bodla2017Soft-NMS} to process all predictions to suppress redundant proposals.

\section{Experiments}

\subsection{Datasets}
To verify the effectiveness of our proposed method, we conduct experiments on two benchmark datasets.

{\bf THUMOS14 \cite{Jiang2014THUMOS14}} dataset contains 1010 validation videos and 1574 testing videos with 101 action categories. We follow the \cite{Jiang2014THUMOS14}, 200 untrimmed videos in the validation set and 213 untrimmed videos in the test set are used for training and testing, respectively. These videos contain 20 categories labeled for temporal action localization. Each video has more than 15 action annotations.

{\bf ActivityNet1.3  \cite{Heilbron2015ActivityNet}} dataset contains 19,994 untrimmed videos with 200 action categories. We follow the setting in \cite{Heilbron2015ActivityNet} and divide the dataset into training, testing and validation with a ratio of 2:1:1. There are around 1.5 action instances for each video.

\begin{table*}[!htbp]
\centering
\caption{Performance comparison with state-of-the-art methods on THUMOS14 dataset measured by mAP with different tIoU
thresholds. The bold numbers represent the best performance.}
\renewcommand{\arraystretch}{1.2}
\setlength{\tabcolsep}{1.8mm}
\begin{tabular}{c|c|c|ccccccc|c|c}
\hline
{Type} & {Methods}  & {Backbone} & {0.1} & {0.2} & {0.3} & {0.4} & {0.5} & {0.6} & {0.7}  & Avg \{0.1:0.5\} & Avg \{0.3:0.7\}\\

\hline

{Anchor-based}
&SSAD \cite{Lin2017SSAD} & TS &50.1 &47.8 &43.0 &35.0 &24.6 &- &- &40.1 &- \cr
&TURN \cite{Gao2017TURN} &C3D &54.0 &50.9 &44.1 &34.9 &25.6 &- &- &41.9 &- \cr
&R-C3D  \cite{Xu2017R-C3D}&C3D &54.5 &51.5 &44.8 &35.6 &28.9 &- &- &43.1 &- \cr
&CBR \cite{Gao2017CBR} &TS &60.1 &56.7 &50.1 &41.3 &31.0 &19.1 &9.9 &47.8 &30.3 \cr
&TALNet \cite{Chao2018Rethinking} &I3D &59.8 &57.1 &53.2 &48.5 &42.8 &33.8 &20.8 &52.3 &39.8 \cr
&GTAN \cite{Long2019GTAN} &P3D &69.1 &63.7 &57.8 &47.2 &38.8 &- &- &55.3 &- \cr
&PCG-TAL \cite{Su2021PCG-TAL} &I3D &71.2 &68.9 &65.1 &59.5 &51.2 &- &- &63.2 &- \\
\hline
{Actionness}
&CDC \cite{Shou2017CDC} & - &- &- &40.1 &29.4 &23.3 &13.1 &7.9 &- &22.8 \cr
&SSN \cite{Zhao2017SSN} &TS &60.3 &56.2 &50.6 &40.8 &29.1 &- &- &47.4 &- \cr
&TAG  \cite{Xiong2017TAG}&TS &64.1 &57.7 &48.7 &39.8 &28.2 &- &- &47.7 &- \cr
&BSN \cite{Lin2018BSN} &TS &- &- &53.5 &45.0 &36.9 &28.4 &20.0 &- &36.8 \cr
&BMN \cite{Lin2019BMN} &TS &- &- &56.0 &47.4 &38.8 &29.7 &20.5 &- &38.5 \cr
&DBG \cite{Lin2020DBG} &TS &- &- &57.8 &49.4 &42.8 &33.8 &21.7 &- &41.1 \cr
&GTAD \cite{Xu2020G-TAD} &TS &- &- &54.5 &47.6 &40.2 &30.8 &23.4 &- &39.3 \\
&BSN++ \cite{Su2021BSN++} &TS &- &- &59.9 &49.5 &41.3 &31.9 &22.8 &- &41.1 \cr
&BU-TAL \cite{Zhao2020Bottom-Up} &TS &- &- &53.9 &50.7 &45.4 &38.0 &28.5 &- &43.3 \cr
&TCA-Net \cite{Qing2021TCA-Net} &TS &- &- &60.6 &53.2 &44.6 &36.8 &26.7 &- &44.4 \cr
&RTD-Action \cite{Tan2021RTD-Net} &TS &- &- &68.3 &62.3 &51.9 &38.8 &23.7 &- &49.0 \cr
&RCL \cite{Wang2022RCL} &TS &- &- &70.1 &62.3 &52.9 &42.7 &30.7 &- &51.7 \cr
&DCAN \cite{Chen2022dcan} &TS &- &- &68.2 &62.7 &54.1 &43.9 &32.6 &- &52.3 \cr
\hline  
{Others}   
&SCNN \cite{Shou2016SCNN} &- &47.7 &43.5 &36.3 &28.7 &19.0 &- &- &19.0 &- \cr
&GTAD+PGCN \cite{Xu2020G-TAD} &TS &- &- &66.4 &60.4 &51.6 &37.6 &22.9 &- &47.8 \cr
&ContextLoc \cite{Zhu2021Enriching} &I3D &- &- &68.3 &63.8 &54.3 &41.8 &26.2 &- &50.9 \cr
&Sub-Action \cite{Wang2022Exploring} &I3D &66.1 &60 &52.3 &43.2 &32.9 &- &- &50.9 &- \cr
&VSGN \cite{Zhao2021VSGN} &TS &- &- &66.7 &60.4 &52.4 &41.0 &30.4 &- &50.2 \cr
\hline
{Anchor-free}   
&A2Net \cite{Yang2020Revisiting} &I3D &61.1 &60.2 &58.6 &54.1 &45.5 &32.5 &17.2 &55.9 &41.6 \cr
&SRF-Net \cite{Ou2022SRFNet} &C3D &- &- &56.5 &50.7 &44.8 &33.0 &20.9 &- &41.2 \cr
&AFSD \cite{Lin2021AFSD} &I3D &- &- &67.3 &62.4 &55.5 &43.7 &31.1 &- &52.0 \cr
&\textbf{TAL-MTS} $\left(Ours\right)$ &I3D &\textbf{75.3} &\textbf{73.8} &\textbf{70.5} &\textbf{65.0} &\textbf{56.9} &\textbf{46.0} &\textbf{32.7} &\textbf{68.3} &\textbf{54.2} \cr
\hline
\end{tabular}
\label{table1}
\end{table*}

\subsection{Experimental Settings}
\textbf{Parameters}. In our experiments, we follow the experimental setup of \cite{Lin2021AFSD}. On the THUMOS14 dataset, we sample RGB and optical flow frames using a frame rate of 10 frames per second (fps) and split the video into clips. For each clip, we set its length $T$ to 256 frames, and adjacent clips will have temporal overlap, which is set to 30 in training and 128 in testing. On the ActivityNet1.3 dataset, the frames are sampled by different fps, we guarantee that the number of each video frame is 768. On both datasets, we set the size of each frame to 96 $\times$ 96 and the size of frame patches $D$ to 24. We also use random crop, and horizontal flipping as the data augmentation during training.

We use Adam \cite{Kingma2015Adam} optimizer for model training and the epochs are set to 25. Besides, the learning rate is ${10}^{-5}$, the weight decay is ${10}^{-3}$, and the batch size is set to 1. The hyperparameters are empirically defined as $\alpha=10$ and $\beta=1$ on THUMOS14 dataset. On the ActivityNet1.3 dataset, $\alpha=1$ and $\beta=1$. In addition, the tIoU threshold in Soft-NMS \cite{Bodla2017Soft-NMS} is set to 0.3 for THUMOS14 and 0.85 for ActivityNet1.3, respectively.

\textbf{Evaluation metrics}. In the TAL task, the mean Average Precision (mAP) is used as the evaluation metric. We report mAP for all experiments. Besides, the tIoU thresholds are [0.1 : 0.1 : 0.7] for THUMOS14 and [0.5 : 0.05 : 0.95] for ActivityNet1.3.

\subsection{Comparison with state-of-the-art methods}
We compare the proposed TAL-MTS method with several state-of-the-art methods, which contain anchor-based, actionness-guided, anchor-free, and other approaches. Table \ref{table1} and Table \ref{table2} report the comparison results on THUMOS14 and ActivityNet1.3 dataset, respectively. Then we discuss the results as follows.

{\bf Results on THUMOS14.} Table \ref{table1} reports the comparison results on the THUMOS14 dataset. For some methods, their tIoU thresholds of 0.1 and 0.2 or 0.6 and 0.7 are not reported. Therefore, we report ${Avg\{0.1:0.5\}}$ and ${Avg\{0.3:0.7\}}$, which are represent the average mAP for all tIoU thresholds $\{0.1:0.1:0.5\}$ and $\{0.3:0.1:0.7\}$, respectively. Our method outperforms strong opponents AFSD, RTD-Action, RCL, DCAN and ContextLoc 
for all thresholds on ${Avg\{0.3:0.7\}}$, and exceeded 2.3\%, 5.2\%, 2.5\%, 1.9\% and 3.3\%, respectively. For the ${Avg\{0.1:0.5\}}$, the TAL-MTS method outperforms PCG-TAL and A2Net by 5.1\% and 12.4\%, respectively. These results show that our performance significantly outperforms the current state-of-the-art methods, and the TAL-MTS method improves from 55.5\% to 56.9\%, when the threshold is 0.5. At the threshold 0.6, the TAL-MTS method exceeds DCAN method 2.1\% , the TAL-MTS method is slightly improved compared with DCAN for thresholds 0.7. Especially when the threshold is 0.4, the TAL-MTS method is more than 2\% higher than state-of-the-art methods. TAL-MTS method is only 0.4\%  higher than the latest RCL method at threshold 0.3.  For thresholds 0.1 and 0.2, the TAL-MTS method is more than 3\% than PCG-TAL and more than 13\% better than A2Net. Although A2Net and AFSD are anchor-free methods, A2Net leverages the advantages of anchor-base and anchor-free and AFSD focuses on refining the boundaries with a saliency-based 
refinement module, which ignores the importance of enough semantics and long-range spatial-temporal information. In TAL-MTS, features of multi-temporal scales with this information are used for TAL. From the above discussion results, we can see that we have the best results for all tIoU thresholds. Therefore, for anchor-based, anchor-free, etc. methods, the TAL-MTS method outperforms these methods by a large margin at all tIoU thresholds.

\begin{table}[!htbp]
\centering
\caption{Performance comparison with state-of-the-art methods on ActivityNet1.3 dataset measured by mAP at different tIoU
thresholds.}
\renewcommand{\arraystretch}{1.2}
\setlength{\tabcolsep}{1.6 mm}
\begin{tabular}{c|c|cccc}
\hline
{Type} & {Methods} & {0.5} & {0.75} & {0.95}  & {Avg} \\

\hline

{Anchor-based}
&R-C3D  \cite{Xu2017R-C3D} &26.8 &- &-  &- \cr
&TALNet \cite{Chao2018Rethinking}  &38.2 &18.3 &1.3  &20.2 \cr
&GTAN \cite{Long2019GTAN}  &52.6 &34.1 &8.9  &34.3 \cr
&PCG-TAL \cite{Su2021PCG-TAL}  &44.3&29.9 &5.5  &28.9 \\
\hline
{Actionness}
&CDC \cite{Shou2017CDC}  &45.3 &26.0 &0.2  &23.8 \cr
&SSN \cite{Zhao2017SSN}  &43.2 &28.7 &5.6   &28.3 \cr
&TAG  \cite{Xiong2017TAG} &41.1 &24.1 &5.0   &24.9 \cr
&BSN \cite{Lin2018BSN}  &46.5&30.0 &8.0  &30.0 \cr
&BMN \cite{Lin2019BMN}  &50.1 &34.8 &8.3  &33.9 \cr
&GTAD \cite{Xu2020G-TAD}  &50.4 &34.6 &9.0  &34.1 \\
&BC-GNN \cite{Bai2020BC-GNN}  &50.6 &34.8 &9.4 &34.3 \cr
&BSN++ \cite{Su2021BSN++}  &51.3 &35.7 &9.0 &34.9 \cr
&BU-TAL \cite{Zhao2020Bottom-Up}  &43.5 &33.9 &\textbf{9.2}&30.1 \cr
&RTD-Action \cite{Tan2021RTD-Net}  &47.2 &30.7 &8.6 &30.8 \cr
&TCA-Net[BMN] \cite{Qing2021TCA-Net}  &52.3 &\textbf{36.7} &6.9 &\textbf{35.5} \cr
\hline
{Others}   
&PGCN \cite{Zeng2019PGCN} &48.3 &33.2 &3.3  &31.1 \cr
&ContextLoc \cite{Zhu2021Enriching}  &\textbf{56.0} &35.2 &3.6 &34.2 \cr
&VSGN \cite{Zhao2021VSGN}  &52.3 &35.2 &8.3 &34.7 \cr
&Sub-Action \cite{Wang2022Exploring}  &37.1 &24.1 &5.8 &24.1 \cr
\hline
{Anchor-free}   
&A2Net \cite{Yang2020Revisiting} &43.6 &28.7 &3.7  &27.8 \cr
&AFSD \cite{Lin2021AFSD}  &52.4 &35.3 &6.5 &34.4 \cr
&\textbf{TAL-MTS} $\left(Ours\right)$  &52.4 &34.7 &6.0 &34.1 \cr
\hline
\end{tabular}
\label{table2}
\end{table}

{\bf Results on ActivityNet1.3.} 
Table \ref{table2} shows the comparison results on the ActivityNet1.3 dataset, where $Avg$ indicates the average mAP for all tIoU thresholds \{0.5:0.05:0.95\}. TAL-MTS method can obtain 
comparable results with other state-of-the-art methods on the ActivityNet1.3 dataset. Because the action instances in the ActivityNet1.3 dataset are long and the scenes where some actions occur are discontinuous, it is very challenging for the anchor-free based methods to detect all the action instances. 
Thus, most TAL methods cannot achieve the best performance on both datasets simultaneously. For example, RTD-Action can achieve better performance on the THUMOS14 dataset and the result is 49.0\%  for ${Avg\{0.3:0.7\}}$, but only 30.8\% on the  ActivityNet1.3 dataset. Similarly, BSN++ achieves 34.9\% on the ActivityNet1.3 dataset, but it is inefficient on the THUMOS14 dataset and the result only is 41.1\% for ${Avg\{0.3:0.7\}}$. The reason for this phenomenon is that the temporal difference between the THUMOS14 dataset and the ActivityNet1.3 dataset is huge. At the same time, it can be seen that TCA-Net achieves the best results on the anet dataset, which is 1.4\% higher than our method, but 9.8\% lower than TAL-MTS on the THUMOS14 dataset. Because the feature extraction method we use on the ActivityNet1.3 dataset is different from TCA-Net, the feature extraction method of TSN \cite{Xiong2016TSN} adopted by TCA-Net often performs better on the ActivityNet1.3 dataset than the I3D \cite{Carreira2017I3D} method. Therefore, some methods can perform well on the THUMOS14 dataset but poorly on the ActivityNet1.3 dataset, and vice versa. From Table \ref{table2}, it can be seen that the actionness-guided methods are more suitable for the ActivityNet1.3 dataset, and they are more likely to obtain better performance. In addition, ContextLoc is a strong opponent on THUMOS14 dataset, which is 3.3\% lower than TAL-MTS for ${Avg\{0.3:0.7\}}$. But on the ActivityNet1.3 dataset, it is only 0.1\% higher than ours. Similar to ContextLoc, Comparing TAL-MTS with GTAD has the same result on the ActivityNet1.3 dataset, but TAL-MTS is 14.9\% higher than GTAD on the THUMOS14 dataset for ${Avg\{0.3:0.7\}}$. For the GTAN, our method is 13\% higher than it on the THUMOS14 dataset for ${Avg\{0.1:0.5\}}$. Yet, TAL-MTS only is 0.2\% lower than GTAN on the ActivityNet1.3 dataset.
Thus, although the proposed TAL-MTS method does not achieve the best results, its performance is still competitive. From the above experimental analysis, it can be concluded that our method has good generalization ability.

{\bf Inference time.} 
We compare the inference speed between TAL-MTS and other methods on the THUMOS14 dataset. Our model is evaluated on a Tesla V100 GPU, its speed of inference is 1824 fps. As shown in Table \ref{table6}, TAL-MTS is fast and the speed of inference reached second place. Our method is slower than AFSD because of two aspects. First, we use a transformer to excavate spatial-temporal information between frames. Next, we divide each frame of the video into patches and pass each patch through the self-attention mechanism. 
These require a higher computational cost. Nevertheless, TAL-MTS is still efficient.

\begin{table}[!htbp]
\centering
\caption{ Comparison of inference time on THUMOS14 dataset.}
\renewcommand{\arraystretch}{1.2}
\setlength{\tabcolsep}{4.4mm}
\begin{tabular}{c|c|c}
\hline
{Model}& GPU & FPS  \cr
\hline
{SCNN}\cite{Shou2016SCNN}& - & 60 \\
{CDC}\cite{Shou2017CDC}& TITAN Xm & 500  \\
{R-C3D}\cite{Xu2017R-C3D}& TITAN Xm & 569 \cr
{R-C3D}\cite{Xu2017R-C3D}& TITAN Xp & 1030 \cr


{PBRNet}\cite{Xiao2020PBRNet}& 1080Ti  & 1488 \cr
{AFSD}\cite{Lin2021AFSD}& 1080Ti  & 3259 \cr
{AFSD}\cite{Lin2021AFSD}& V100  & 4057 \cr
\hline
\textbf{TAL-MTS} $\left(Ours\right)$& V100 & 1824  \\
\hline
\end{tabular}
\label{table6}
\end{table}

\section {Ablation Study}

To verify the effectiveness of each module in our method, we follow the previous methods \cite{Lin2021AFSD, Wang2021RGB, Zeng2019PGCN, Zhao2020Bottom-Up, Ou2022SRFNet, Yang2020Revisiting, Wang2022TVNet} to conduct the ablation study on the THUMOS14 dataset and ActivityNet1.3 dataset. We mainly focus on the following aspects: validity of the RGB and optical-flow streams, effects of RFP, STT, and FSA modules, the effect of patch size for FSA, and the impact of different tIoU thresholds, visualization prediction boundaries, and loss convergence analysis.

\subsection { Validity of the RGB and optical-flow streams.}
We use both the RGB stream and optical-flow stream for TAL and combine them with an average fusion. To further verify the effectiveness of each stream, we conduct experiments on the ActivityNet1.3 dataset and THUMOS14 dataset. From Table \ref{table_anet}, for ActivityNet1.3 dataset, the RGB stream with $Avg$ is 32.5\% and the optical-flow stream with $Avg$ is 32.6\%, but the result of fusion is 34.1\%. We can observe better results for optical-flow stream at tIoU thresholds of 0.5 and 0.75, and better results for RGB stream at tIoU threshold of 0.95. Similarly, for the THUMOS14 dataset, it can be intuitively seen that the a huge difference between single stream and two-stream results. The RGB stream with ${Avg\{0.1:0.7\}}$ is 50.9\% and the optical-flow stream with ${Avg\{0.1:0.7\}}$ is 50.4\%, but the result of fusion is 60.0\%. From the average of three data, we can be seen that the result of the fusion is 9.1\% higher than the RGB stream, and 9.6\% higher than the optical-flow stream. Similarly, we can see from the table that RGB and optical-flow streams are complementary. When the temporal tIoU threshold is low, the RGB stream can get better results, and the tIoU threshold is high, the optical-flow stream can get good results. For example, the RGB stream has good performance, and the tIoU threshold is from 0.1 to 0.4. The optical-flow stream is higher than RGB from 0.5 to 0.7. Meanwhile, the dense optical flow field can provide motion information, which is very important for the acquisition of spatial-temporal information. Thus, the two-stream network is more effective in the two datasets.

\begin{table}[!htbp]
\centering
\caption{ Validity of the RGB and optical flow stream
on ActivityNet1.3 dataset.}
\renewcommand{\arraystretch}{1.2}
\setlength{\tabcolsep}{3.5mm}
\begin{tabular}{c|ccc|c}
\hline
{}  & {0.5} & {0.75} & {0.95}  &Avg   \\

\hline

 RGB  &50.1 &33.0 &\textbf{6.5} &32.5 \cr
 \hline
 Flow   &51.1 &33.4 &4.9  &32.6 \cr
 \hline
Fusion &\textbf{52.4} &\textbf{34.7}  &6.0 &\textbf{34.1} \cr
\hline
\end{tabular}
\label{table_anet}
\end{table}

\begin{table}[!htbp]
\centering
\caption{ Validity of the RGB and optical flow stream
on THUMOS14 dataset.}
\renewcommand{\arraystretch}{1.2}
\setlength{\tabcolsep}{1.0mm}
\begin{tabular}{c|ccccccc|c}
\hline
{}  & {0.1} & {0.2} & {0.3} & {0.4} & {0.5} & {0.6} & {0.7}  &Avg  \{0.1:0.7\} \\

\hline

 RGB  &66.4 &64.7 &61.5 &55.4 &46.9 &37.0 &24.2 &50.9 \cr
 \hline
 Flow   &62.5 &61.0 &58.4 &54.7 &48.6 &39.4 &28.4 &50.4 \cr
 \hline
Fusion &\textbf{75.3} &\textbf{73.8} &\textbf{70.5} &\textbf{65.0} &\textbf{56.9} &\textbf{46.0} &\textbf{32.7} &\textbf{60.0} \cr
\hline
\end{tabular}
\label{table3}
\end{table}

\subsection { Effects of RFP, STT, and FSA modules.}
To verify the effects of our proposed modules, we evaluate these modules and verify the feasibility of module stacking. In Table \ref{table5}, the baseline reaches $52.0\%$ on the THUMOS14 dataset. When we add the RFP to refined feature pyramids with semantics information at different temporal scales, the improvement is $0.6\%$, which shows that the semantics information of features is indispensable for TAL tasks.
Besides, we propose an STT to fully excavate spatial-temporal information and generate features with long-range dependencies in the video, which improves the performance by $0.8\%$. Then, by combining RFP and STT, the features of multi-temporal scales are gained, and the average mAP is further improved by $1.1\%$. This shows the effectiveness of the proposed two modules, and features of multi-temporal scales are very important for TAL-MTS.
Additionally, we use an FSA module for distinguishing foreground and background, which captures the foreground information inside each frame in the video and reduces noises, and further improves the localization and classification of action instances. By fusing all three modules, the proposed TAL-MTS method achieves the best performance ($54.2\%$) on the THUMOS14 dataset, which outperforms the baseline by $2.2\%$. Based on the above analysis, it can be seen that our proposed three modules are very effective for TAL tasks.

\begin{table}[!htbp]
\centering
\caption{ Benefits of RFP, STT and FSA on THUMOS14.}
\renewcommand{\arraystretch}{1.2}
\setlength{\tabcolsep}{3.2mm}
\begin{tabular}{c|ccccc}

\hline
{baseline} & \Checkmark  & \Checkmark & \Checkmark & \Checkmark & \Checkmark\cr
{RFP }&  & \Checkmark & & \Checkmark & \Checkmark \\
{STT }&  & & \Checkmark & \Checkmark  & \Checkmark\\
{FSA}& & & & & \Checkmark\cr
\hline
{Avg \{0.3:0.7\}}& 52.0 & 52.6 &52.8 & 53.1 & \bf{54.2} \cr
\hline
\end{tabular}
\label{table5}
\end{table}

\begin{figure}[htbp]
\centering 
\includegraphics[width=0.85\linewidth,height=0.25\textwidth]{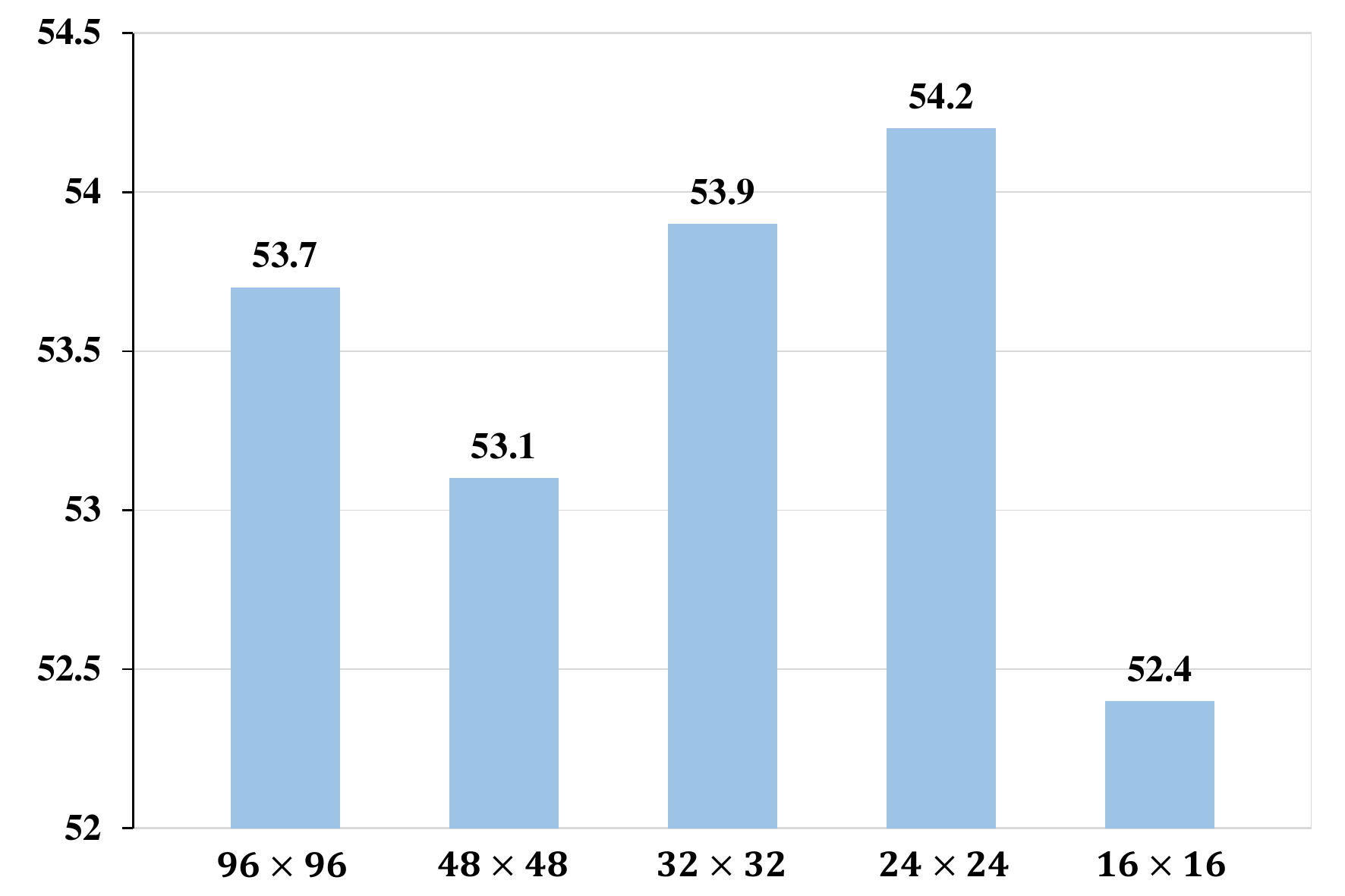}

\caption{The average mAP for different patch sizes on the THUMOS14
dataset, where the x-coordinate indicates the patch size,
and the y-coordinate denotes $Avg\{0.3:0.7\}$.}
\label{fig_patch}
\end{figure}

\begin{figure*}[htbp]
\centering 
\includegraphics[width=0.95\linewidth, height=0.55\textwidth]{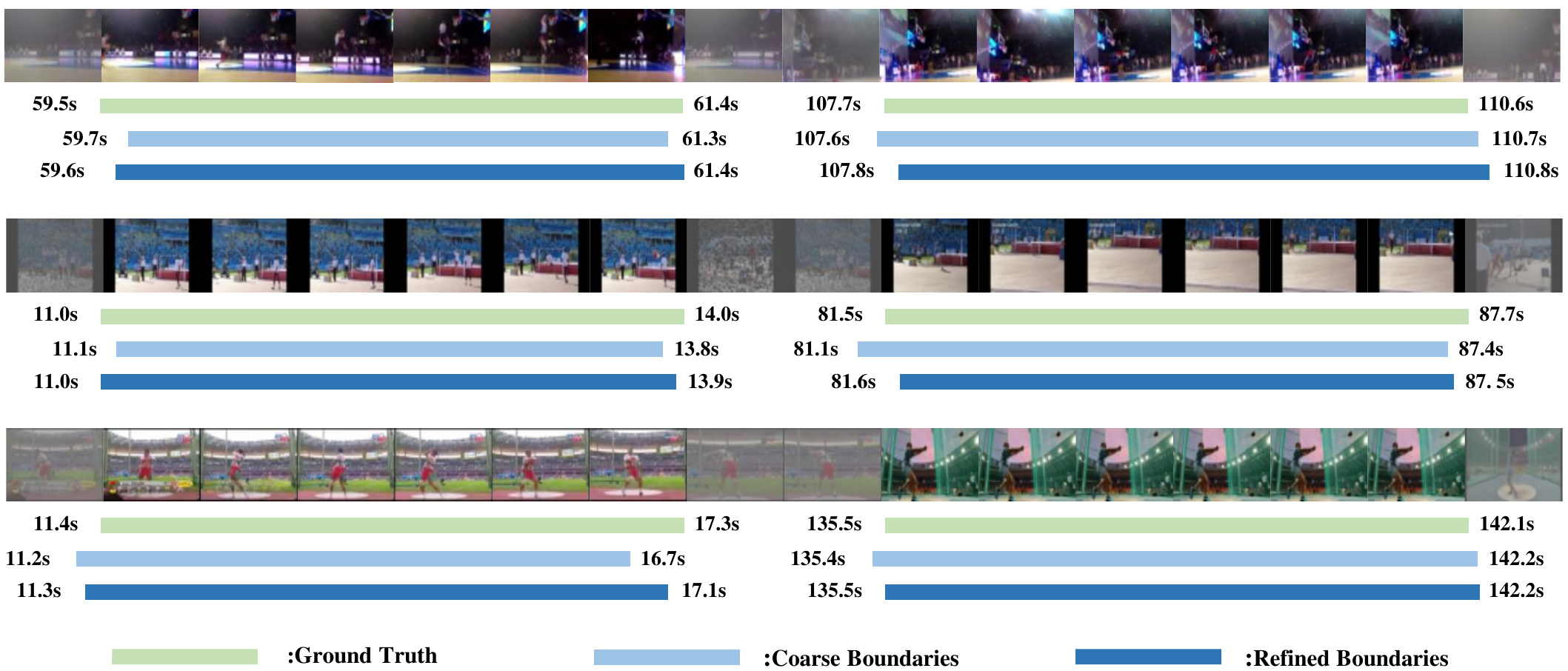}

\caption{Visualization examples of the boundaries predicted by the proposed TAL-MTS on THUMOS14 dataset.
}
\label{fig_visual}
\end{figure*}

\subsection {The effect of patch size for FSA.} 
Because the dividing frame strategy can process each patch, more details can be obtained, which is very important for maximizing the separation of foreground action and background noise. Therefore, we divide all frames of the whole video, and our FSA models the patch of $D \times D$. Because the size of the video frame patches has an impact on separating foreground and background noise, in this section, we discuss the effect of the patch of different sizes on TAL-MTS. As shown in Figure \ref{fig_patch}, A patch size of $96 \times 96$ means that we directly perform self-attention on the original video frame, and we can find that the second-best result is obtained. Because this does not take into account the fine-grained information within each video frame, it can not separate the foreground action and background noise to the greatest extent. On the contrary, when we divide each video frame, we can capture fine-grained information and greatly reduce the impact of background noise. When patch size is set to $24 \times 24$, TAL-MTS has the best results on average. From the figure, we can see that different patch sizes have different 
results. Thus, we test the patch size from large to small (Note that the patch size must be divided by the size of the video frame), and we find that when the patch size increased from $48 \times 48$ to $24 \times 24$, the effect was steadily improving. But the effect becomes worse when the patch size is $16 \times 16$. Because 
the distinction between foreground action and background is not obvious enough in some patches, thus, the features of these patches will be mistaken for the background to affect the separation of foreground action and background. Thus, a suitable patch size can enhance foreground action information and better separate foreground actions and background noises. Finally, the patch size is set to $24 \times 24$ and the ${Avg\{0.3:0.7\}}$ is 54.2\%.

\subsection {Impact of different tIoU thresholds.}
For the TAL-MTS method to achieve the best performance, we follow the approach of \cite{Yang2020Revisiting} and \cite{Wang2022TVNet}, and discuss the impact of different tIoU thresholds on the THUMOS14 dataset. As can be seen from Table \ref{table4}, when the tIoU threshold is 0.2, mAPs of 0.3 and 0.4 achieve the best performance. When the tIoU threshold is 0.5, mAPs of 0.6 and 0.7 achieve the best performance. When the tIoU threshold is 0.3, only mAP of 0.5 achieves the best performance, but the Avg which is 0.15\% higher than tIoU threshold of 0.2 and also 0.21\% higher than tIoU threshold of 0.5 achieves the best result. Although other tIoU thresholds achieve the best results at different mAPs, we still set the tIoU threshold to 0.3 because it achieves the best performance on average. Setting the tIoU threshold to 0.3 achieves as good performance as possible on all mAPs, which makes the fitting better of the TAL-MTS method better.

\begin{table}[!htbp]

\centering
\caption{ The impact of different tIoU threshold in Soft-NMS on THUMOS14 dataset.}
\renewcommand{\arraystretch}{1.2}
\setlength{\tabcolsep}{1.3mm}
\begin{tabular}{c|ccccc|c}
\hline

\multirow{2}*{tIoU threshold}& {} & {} & {mAPs} & {} & {}&\multirow{2}* {Avg} \\ \cline{2-6}
{}& {0.3} & {0.4} & {0.5} & {0.6} & {0.7}  &{} \\

\hline

 0.2    &\textbf{70.80} &\textbf{65.12} &56.73 &45.63 &32.10 &54.08\\


0.3  &70.50 &65.04 &\textbf{56.89} &46.03 &32.71 &\textbf{54.23} \\
0.4     &70.06 &64.74 &56.72 &46.16 &33.11 &54.16\\
0.5     &69.56 &64.37 &56.53 &\textbf{46.19} &\textbf{33.29} &53.99\\
\hline
\end{tabular}
\label{table4}
\end{table}

\begin{figure}[htbp]
\centering 
\includegraphics[width=0.9\linewidth,height=0.35\textwidth]{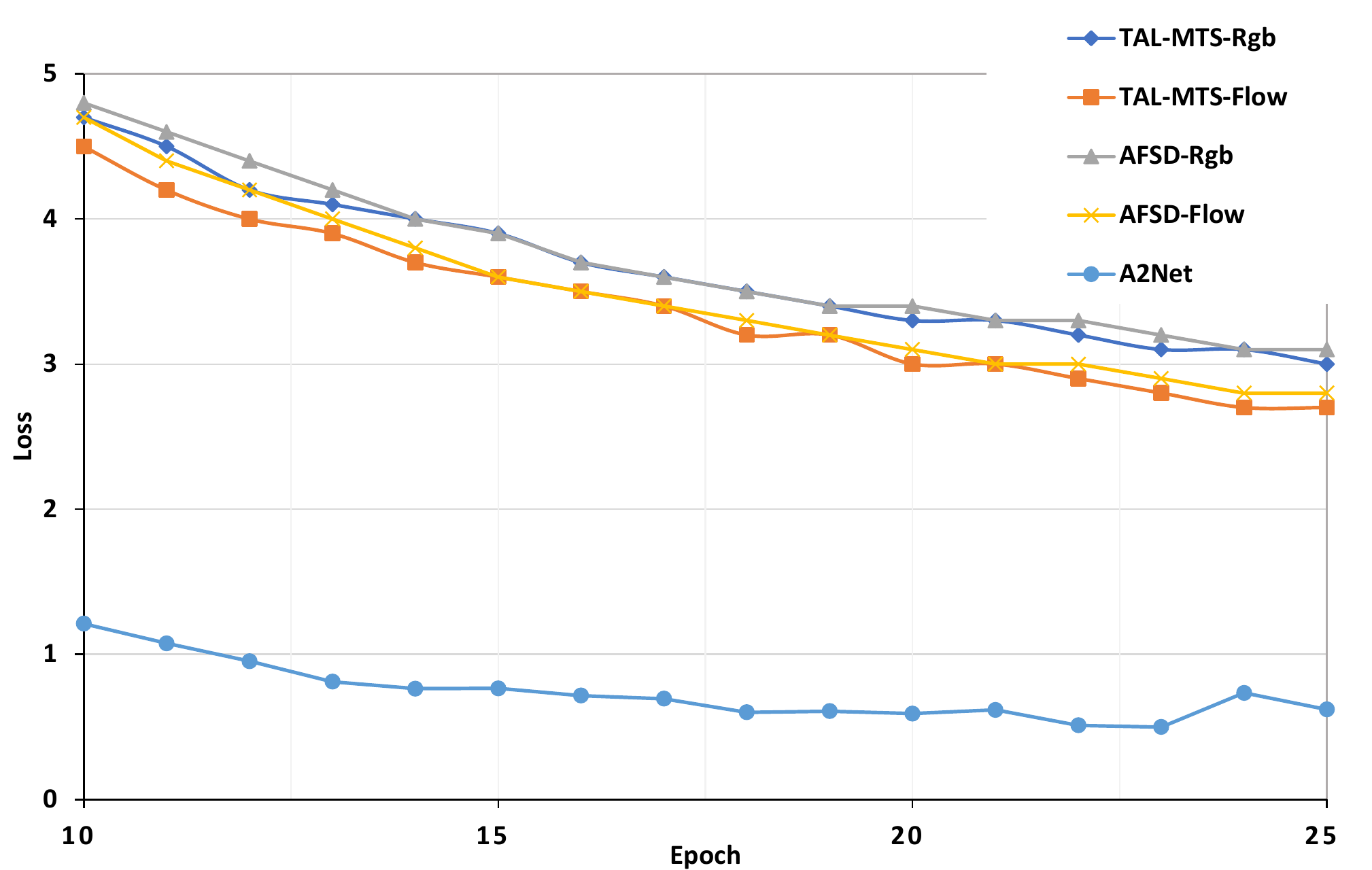}

\caption{Loss convergence curves of TAL-MTS on the THUMOS14
dataset, where the x-coordinate denotes the number of epochs,
and the y-coordinate indicates the loss values.
}
\label{fig_loss}
\end{figure}

\subsection {Visualization prediction boundaries.} 

In this section, we visualize the comparison between the predicted boundaries of TAL-MTS and the ground truth on the THUMOS14 dataset. Some of the results are shown in Figure \ref{fig_visual}. The coarse boundaries represent the results without the FSA refinement, and the refined boundaries represent the results after the FSA refinement. We can be seen that although the coarse boundaries are very close to the ground truth, the refined boundaries are more accurate after the FSA refinement. This shows that the FSA model is very effective for TAL-MTS to refine results.

\subsection {Loss convergence analysis.} 

To further demonstrate the effectiveness of the TAL-MTS method, in this section, we analyze the loss convergence of TAL-MTS on the THUMOS14 dataset and compare it with other anchor-free state-of-the-art methods, such as AFSD \cite{Lin2021AFSD} and A2Net \cite{Yang2020Revisiting}. As shown in Figure \ref{fig_loss}, we give the RGB stream and optical-flow stream loss curves of AFSD  and TAL-MTS, and also give the loss curve of A2Net. As can be seen from these curves, the TAL-MTS method Whether in the RGB stream or optical-flow stream can converge in $20-25$ epochs, so the convergence speed is very fast. Compared with AFSD, our method converges faster in both RGB stream, and optical-flow stream, thus, further proving that our method is effective.

\section{Conclusion}

This paper proposes a novel temporal action localization method with refined features of multi-temporal scales. First, RFP is proposed to obtain refined feature pyramids with 
stronger semantics information. Then, STT is designed to fully excavate the long-range dependencies of video frames. By combining the two modules, features of multi-temporal scales can be generated.
Finally, we use an FSA module to capture the foreground information of each frame, which can reduce the influence of background noises, and further refine the results of localization and classification. Furthermore, we jointly optimize the three models in the framework. Extensive experiments show that we achieve significant results on the THUMOS14 dataset and comparable results on the ActivityNet1.3 dataset. The results show that TAL-MTS is an effective method in the temporal action localization task.

\ifCLASSOPTIONcaptionsoff
  \newpage
\fi

\normalem

\bibliographystyle{plain}
 \bibliography{mybib}

%




\end{document}